\documentclass[conference]{IEEEtran}
\IEEEoverridecommandlockouts
\usepackage{cite}
\usepackage{import}
\usepackage{amsmath,amssymb,amsfonts}
\usepackage{algorithmic}
\usepackage{graphicx}
\usepackage{textcomp}
\usepackage{xcolor}
\usepackage{hyperref}
\usepackage{gensymb}
\usepackage[font=footnotesize]{caption}

\def\BibTeX{{\rm B\kern-.05em{\sc i\kern-.025em b}\kern-.08em
    T\kern-.1667em\lower.7ex\hbox{E}\kern-.125emX}}
\begin{document}

\title{Toward Data-Driven Surrogates of the Solar Wind with Spherical Fourier Neural Operator}

\author{\IEEEauthorblockN{Reza Mansouri}
\IEEEauthorblockA{Georgia State University\\ Atlanta, GA, USA\\
rmansouri1@student.gsu.edu}
\and
\IEEEauthorblockN{Dustin J. Kempton}
\IEEEauthorblockA{Georgia State University\\ Atlanta, GA, USA\\
dkempton1@cs.gsu.edu}
\and
\IEEEauthorblockN{Pete Riley}
\IEEEauthorblockA{Predictive Science Inc.\\ San Diego, CA, USA\\
pete@predsci.com}
\and
\IEEEauthorblockN{Rafal A. Angryk}
\IEEEauthorblockA{Georgia State University\\ Atlanta, GA, USA\\
rangryk@gsu.edu}
}

\maketitle

\begin{abstract}

The solar wind, a continuous stream of charged particles from the Sun’s corona, shapes the heliosphere and impacts space systems near Earth. Variations such as high-speed streams and coronal mass ejections can disrupt satellites, power grids, and communications, making accurate modeling essential for space weather forecasting. While 3D magnetohydrodynamic (MHD) models are used to simulate and investigate these variations in the solar wind, they tend to be computationally expensive, limiting their usefulness in investigating the impacts of boundary condition uncertainty. In this work, we develop a surrogate for steady state solar wind modeling, using a Spherical Fourier Neural Operator (SFNO). We compare our model to a previously developed numerical surrogate for this task called  HUX, and we show that the SFNO achieves comparable or better performance across several metrics. Though HUX retains advantages in physical smoothness, this underscores the need for improved evaluation criteria rather than a flaw in SFNO. As a flexible and trainable approach, SFNO enables efficient real-time forecasting and can improve with more data. The source code and more visual results are available at \href{https://github.com/rezmansouri/solarwind-sfno-velocity}{\texttt{github.com/rezmansouri/solarwind-sfno-velocity}}.


\end{abstract}


\section{Introduction}


The solar wind is a continuous stream of charged particles, primarily protons and electrons, expelled from the Sun’s corona. Driven by thermal and magnetic pressure, it escapes along open magnetic field lines, particularly from coronal holes, and carries solar magnetic fields outward at speeds exceeding a million miles per hour. This flow forms the heliosphere, a vast plasma-filled region extending beyond the planets and structured by the interplanetary magnetic field \cite{nasasolarwind}. Variations in the solar wind, including fast and slow streams and explosive events like coronal mass ejections (CMEs), can disrupt the near-Earth space environment. Such disturbances trigger geomagnetic storms that threaten satellites, navigation systems, and power infrastructure. Accurate solar wind prediction is therefore critical, especially during extreme events such as the 1859 Carrington event \cite{nrc2009, carrington1859description}.

\begin{figure}[!t]
    \centering
    \includegraphics[width=1\linewidth]{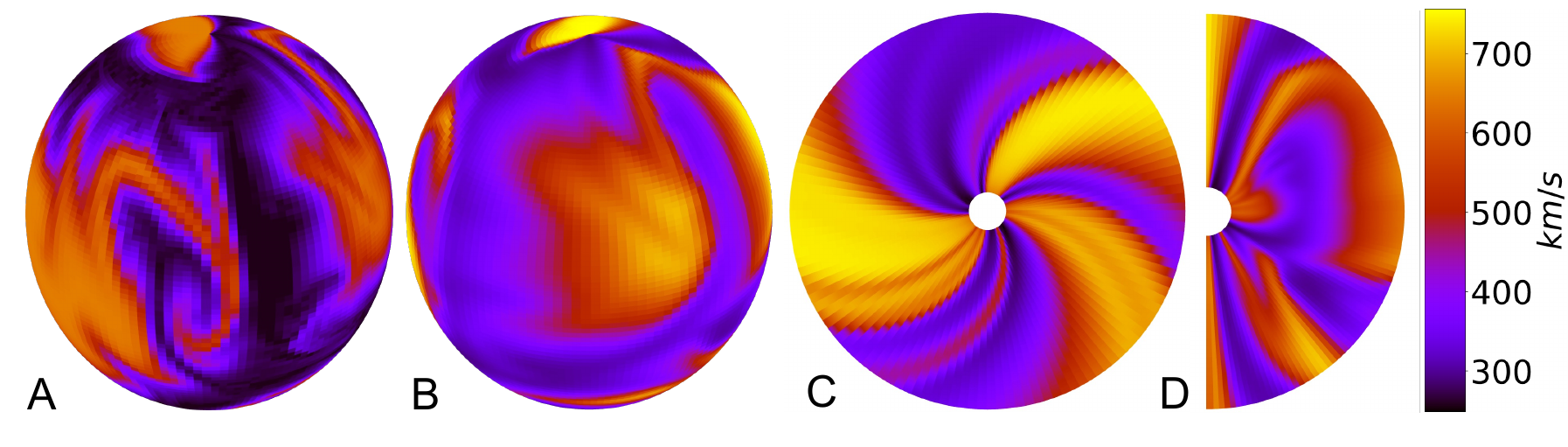}
    \caption{
    Radial velocity from a MAS simulation of Carrington Rotation 2293 (1 Jan–3 Feb 2025). (A) and (B) show velocity at the inner boundary (\(30\,R_\odot\)) and at 1~AU (\(\sim236\,R_\odot\)), respectively. (C) and (D) show slices across all radii at \(90^\circ\) latitude and \(0^\circ\) longitude. The spiral in (C) reflects solar rotation and shapes heliospheric structure. A and B represent spherical surfaces and are not drawn to scale.
    }

    \label{fig:mas_1}
    
\end{figure}

Several numerical surrogates exist for modeling the evolution of the solar wind from the Sun to 1 Astronomical Unit (AU), each balancing different trade-offs between simplicity, speed, and physical fidelity. These include the ballistic extrapolation method \cite{nolte1973, neugebauer1998}, the Arge–Pizzo kinematic evolution model \cite{argepizzo}, the 1-D upwind scheme \cite{riley2011}, the HUX (Heliospheric Upwinding eXtrapolation) method \cite{hux1}, and fully resolved 3D magnetohydrodynamic (MHD) models \cite{mas}. While each model has its use case depending on computational constraints and the level of detail required, the MHD-based MAS (Magnetohydrodynamics Algorithm outside a Sphere) model stands out as the most physically complete and extensively validated approach (see fig.~\ref{fig:mas_1}). It solves the full resistive MHD equations in three dimensions and provides the highest fidelity results for solar wind propagation and stream interaction structure.

The development of machine learning (ML) based surrogates is crucial for modeling the solar wind, because they provide a computationally efficient substitute for traditional first-principles MHD solvers, which are often too slow to meet the demands of operational forecasting. These conventional solvers discretize space to approximate complex PDEs, making high-resolution simulations of the solar wind and CME propagation time-consuming and resource-intensive. In contrast, ML surrogates prioritize reproducing the system's dynamics accurately enough for forecasting and uncertainty quantification, without the computational burden of solving the full physical equations each time \cite{gramacy2020surrogates}. This is especially beneficial when exploring high-dimensional parameter spaces, such as those involved in estimating CME time of arrival and impact, where traditional ensemble modeling becomes intractable \cite{mays2015:ensemble, riley2012:geoeffectiveness}. Since collecting solar wind data is expensive and experimental manipulation is impossible, ML surrogates offer a faster, cheaper, and more scalable path to predictive modeling.

In this work, we propose an ML surrogate built with the Spherical Fourier Neural Operator (SFNO)~\cite{bonev:sfno}, trained to learn the radial velocity field \(v_r(\phi, \theta, r)\) using MAS MHD simulations as the ground truth. Section~\ref{sec:background} reviews related work on solar wind and data-driven surrogate models developed for similar physical systems. Section~\ref{sec:methods} outlines our methodology, including the dataset, modeling pipeline, training, and the evaluation metrics. Finally, section~\ref{sec:results} presents the results and provides a discussion of model fidelity, generalization ability, and relevance to operational heliophysics applications.

\section{Background}
\label{sec:background}

Recent efforts in data-driven solar wind forecasting have explored deep learning models that use either temporal or image-based inputs to improve prediction accuracy. WindNet~\cite{upendran:windnet} predicts daily average solar wind speed using EUV images of the solar corona, notably the 193~\AA{} and 211~\AA{} channels, and learns meaningful features like activation near coronal holes. The Solar Wind Attention Network (SWAN)~\cite{cobos:swan} takes an image-free, autoregressive approach using an attention-based encoder–decoder architecture, achieving lower RMSE than WindNet and earlier CNN-based models.

Unlike models that predict spatial averages, our goal is to learn the full spatial evolution of the solar wind across the heliosphere. This requires mapping between physical states in a way that respects the problem’s symmetries and underlying physics. As standard deep learning models fall short, we adopt an operator learning approach that works in the spectral domain and better aligns with the nature of solar wind propagation.

Operator learning~\cite{operator} has emerged as a powerful framework for modeling mappings between infinite-dimensional function spaces, making it especially suited for problems governed by underlying physical laws. The Fourier Neural Operator (FNO)~\cite{li:fno} introduces learning mappings between function spaces through parameterizing the integral kernel in Fourier space, enabling both expressiveness and efficiency. It achieves state-of-the-art accuracy on PDE benchmarks like Burgers’ equation and Navier–Stokes, and is the first learning-based method to model turbulent flows with zero-shot super-resolution. As an example, FourCastNet\cite{pathak:fourcastnet} is a global data-driven weather forecasting model that uses a Fourier forecast neural network to generate accurate short to medium-range predictions. It excels at forecasting high-resolution variables like precipitation, matching the accuracy of the state-of-the-art forecasting systems.

Spherical Fourier Neural Operators (SFNO)~\cite{bonev:sfno} extend FNOs to spherical domains by using the Spherical Harmonic Transform (SHT) in place of the FFT. This ensures rotational equivariance and improved stability, making SFNOs well-suited for learning spatio-temporal operators on spherical geometries.  

As a baseline solver, we adopt the Heliospheric Upwind eXtrapolation (HUX) model~\cite{hux1}, a reduced-physics model for the solar wind’s radial velocity \(v_r\) beyond 30 solar radii (\(R_\odot\)). HUX simplifies the fluid momentum equation by assuming radial flow and neglecting pressure and gravity, enabling efficient propagation using upwind finite-difference schemes. The model can be run in forward mode (HUX-f), from \(30\,R_\odot\) to 1~AU, or in backward mode (HUX-b), tracing solar wind radial velocity back from 1~AU to \(30\,R_\odot\). In this work, we use HUX-f as our numerical surrogate baseline and compare its performance to our data-driven SFNO model. Section~\ref{sec:methods} describes the modeling pipeline and experimental setup in detail.

\section{Methods}
\label{sec:methods}

    
    


\subsection{Dataset}
Our dataset consists of 616 Carrington rotations spanning from 19 February 1975 to 1 January 2025, covering more than four full solar cycles. These data were compiled from MAS simulations based on boundary conditions at $30\ R_\odot$ produced using observations from three different instruments: the Kitt Peak Observatory (KPO), the {SOHO}/Michelson Doppler Imager (MDI), and the {SDO}/Helioseismic and Magnetic Imager (HMI).



\begin{figure}[!b]
    \vspace{-20pt}
    \centering    \includegraphics[width=1\linewidth]{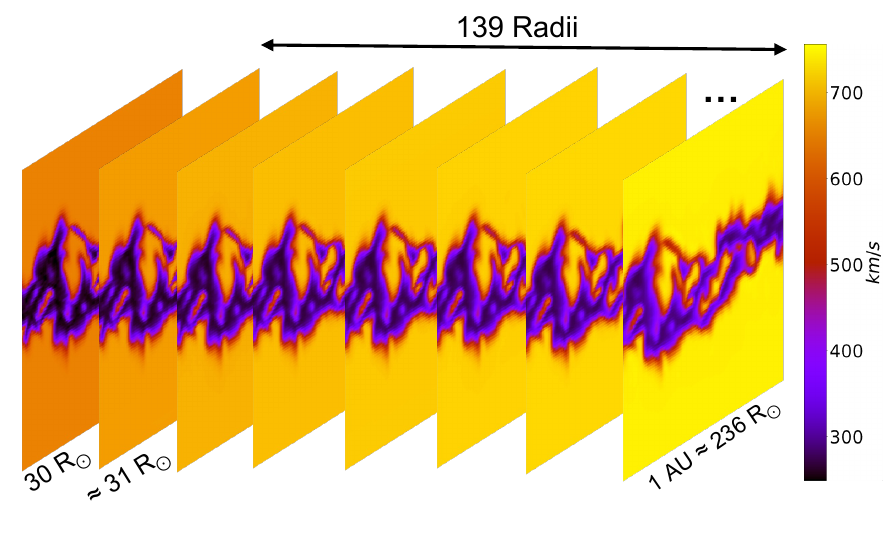}
    \vspace{-20pt}
\caption{
Equiangular projections of solar wind radial velocity in one instance of our dataset (MAS simulation of Carrington Rotation 2234, from 11 August 2020 to 7 September 2020). The first radius \(r_0 = 30\,R_\odot\) serves as the input, and the remaining radii constitute the ground truth.
}
    \label{fig:mas_2}
\end{figure}

With a total of {1,069} samples from different instruments and across the Carrington rotations, each sample is a three-dimensional cube defined on a medium-resolution grid of $(140, 111, 128)$, corresponding to radius, latitude, and longitude respectively. Corresponding with the HUX-f method, in our setup, the first radial slice ($r_0$) of each cube is used as the input, representing the velocity field at the inner boundary of the heliosphere simulation. The remaining 139 radial layers ($r_1$ through $r_{139}$) form the ground truth, capturing the solar velocity structure deeper into the heliosphere (see fig.~\ref{fig:mas_2}).

\subsection{Modeling Pipeline}

Our model is the Spherical Fourier Neural Operator (SFNO)~\cite{bonev:sfno}, a generalization of the Fourier Neural Operator (FNO)~\cite{li:fno} to spherical domains. Unlike FNOs that rely on the FFT and suffer from distortions on the sphere, SFNO uses the Spherical Harmonic Transform (SHT) to respect the symmetries of the domain and ensure rotational equivariance and stability. The architecture consists of an encoder MLP, several SFNO blocks combining pointwise MLPs and spectral convolutions, and a decoder MLP, with position embeddings and skip connections to support stable autoregressive rollout. Moreover, SFNO is grid-invariant, enabling use across different spherical resolutions and meshes.

In our implementation, we use 110 modes in latitude, the maximum permitted by polynomial quadrature rules based on Gauss–Legendre sampling. For the longitudinal direction, which is periodic over \([0, 2\pi]\), we use 64 modes, which is the maximum allowed under the Shannon–Nyquist sampling theorem. The model takes a single input channel corresponding to the solar wind radial velocity at the inner boundary (\(r_0 = 30\,R_\odot\)), and predicts 139 output channels spanning from \(r_1 \approx 31\,R_\odot\) to 1~AU. No tensor factorization is applied in the SFNO layers, allowing for maximum learning capacity. The number of SFNO layers and the number of hidden channels per layer are treated as tunable hyperparameters.

\subsection{Evaluation metrics}

\begin{figure}[!h]
    \centering
    \includegraphics[width=1\linewidth]{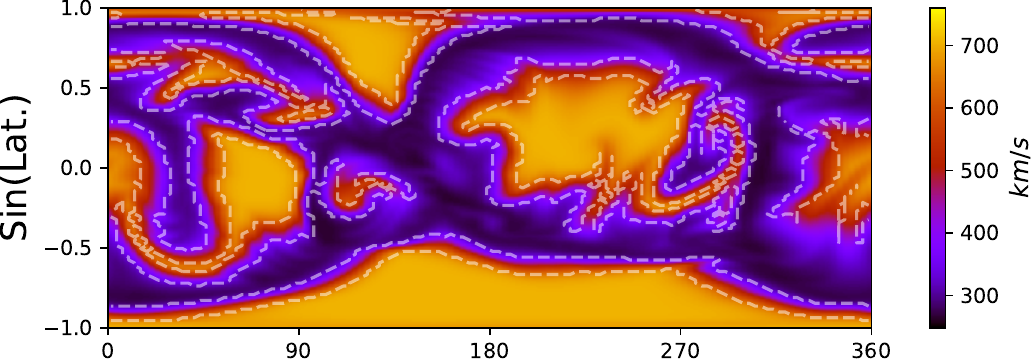}
    \caption{
    Solar wind radial velocity at $R\approx49\ R\odot$ for Carrington Rotation 2293 from the MHD solution. The edge regions, detected using a Sobel filter, are outlined with dashed lines.
}
\vspace{-10pt}
    \label{fig:edges}
\end{figure}

We evaluate model performance using several complementary metrics. Mean Squared Error (MSE), a standard choice for regression tasks, is used during cross-validation to select the best hyperparameters. To emphasize accurate prediction of high-gradient regions, where dynamics such as fast wind overtaking slow wind can generate shocks~\cite{riley:tilts, mas}, we also introduce an edge-specific MSE. This variant computes the MSE only over edge regions identified in radial slices using a $3\times3$ Sobel filter (see Fig.~\ref{fig:edges}). For the final evaluation, we report additional metrics including Earth Mover’s Distance (Wasserstein distance), Multiscale Structural Similarity (MS-SSIM)~\cite{msssim}, Peak Signal-to-Noise Ratio (PSNR), and Anomaly Correlation Coefficient (ACC). The ACC requires a climatology, which we construct by averaging the training data of each task across all samples to obtain a climatology data cube.

\subsection{Training Strategy}

\subsubsection{Loss Function and Optimization}
The model is trained using a layer-wise two-dimensional \(L_{2}\) loss, defined as
\begin{align}
\mathcal{L}_{2}^{(2\mathrm{D})}
&= \frac{1}{BC} \sum^{B\times C}_{b,c}
\left(
\sum^{H\times W}_{i,j}
\lvert y_{bcij} - \hat{y}_{bcij} \rvert^2
\right)^{1/2}
\label{eq:l2_2d_short}
\end{align}
where \(B\) is the batch size, \(C\) the number of output channels (number of radii), and \(H \times W\) the latitude–longitude grid dimensions, and $y$ and $\hat{y}$ are the ground truth and the prediction respectively. We use the Adam optimizer with a fixed learning rate of \(8\times10^{-4}\), training with a batch-size of 32. The model checkpoint with the lowest validation loss is selected as the final model.

\subsubsection{Cross-Validation and Training}
To determine the optimal architecture, we perform cross-validation for 150 epochs using MSE as the selection metric. Each training split is min-max normalized to $[0,1]$, and predictions on validation/test splits are rescaled using the training split’s min and max. We explore models with 4 and 8 SFNO layers and hidden channel sizes of 64, 128, and 256, balancing predictive performance against overfitting risk. Once the best configuration is identified, the model is re-trained from scratch on the full training set for 200 epochs and evaluated on the holdout test set. Table~\ref{tab:dataset} summarizes the dataset split used for this process.

\begin{table}[h!]
\caption{Carrington rotation ranges, corresponding date spans, and instance counts used for training, cross-validation, and testing.}
\centering
\resizebox{\columnwidth}{!}{%
\begin{tabular}{l|l|l|l}
\textbf{Phase} & \textbf{CR \#} & \textbf{Date Range} & \# \textbf{Instances} \\
\hline
Training \& CV & 1625--2169 & 19 Feb. 1975 -- 4 Oct. 2015 & 697 \\
Testing & 2170--2293 & 31 Oct. 2015 -- 1 Jan. 2025 & 372 \\
\end{tabular}%
}
\label{tab:dataset}
\end{table}

\subsubsection{System Specifications}
All models are implemented in Python 3.10.12 using PyTorch version 2.2.1 and the NeuralOperator package~\cite{neuralop1, neuralop2} version 1.0.2. Experiments were conducted on a system running Ubuntu 22.04.3 with an NVIDIA A40 48~GB GPU and approximately 500~GB of RAM. Since HUX is implemented as a CPU-only solver, we also report benchmark results of SFNO inference on CPU for a fair comparison.

\section{Results and Discussion}
\label{sec:results}

\begin{figure}[!b]
    \centering
    \includegraphics[width=1\linewidth]{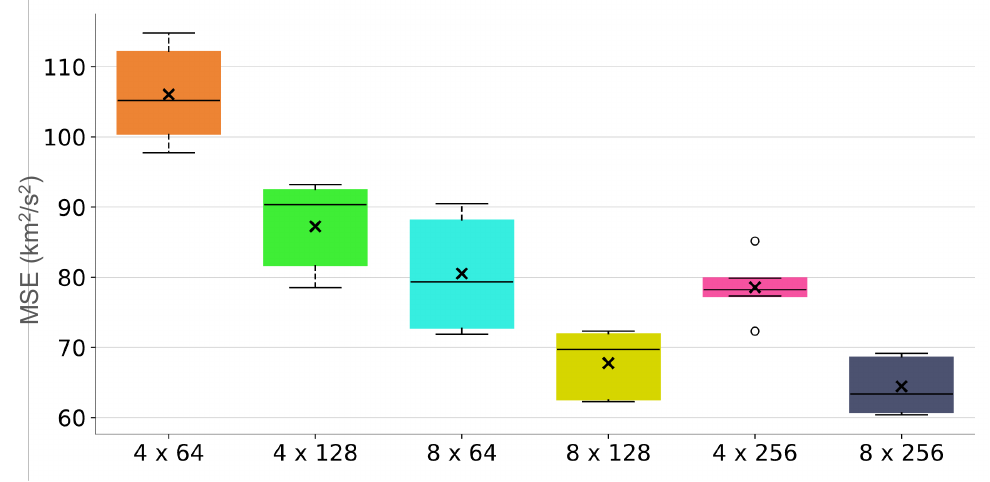}
    \caption{
Cross-validation results showing the MSE across five folds for SFNO architectures with varying depths and hidden channel sizes; for example, $4\times64$ denotes a model with 4 layers and 64 hidden channels.
}
    \label{fig:mse_cv}
\end{figure}

\begin{figure}
    \centering
    \includegraphics[width=1\linewidth]{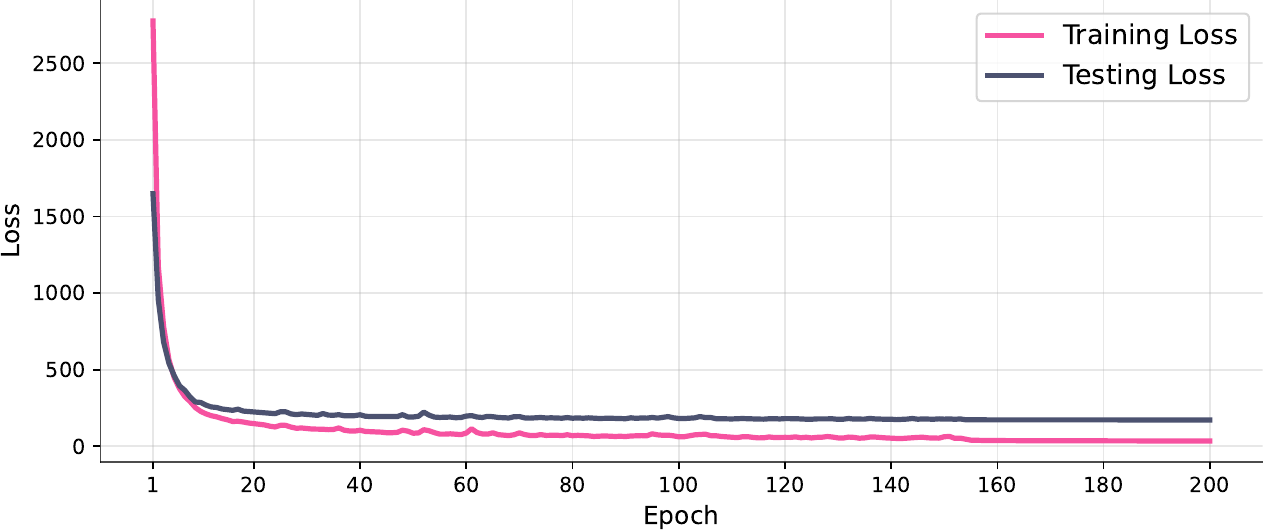}
    \caption{Training and test loss curves for the optimal SFNO configuration ($8\times256$). Both losses decrease consistently, with convergence reached after about the $160^{th}$ epoch, indicating stable generalization to the held-out test set.}
\label{fig:loss}
\vspace{-10pt}
\end{figure}

\begin{figure}[!h]
    \centering
    \includegraphics[width=1\linewidth]{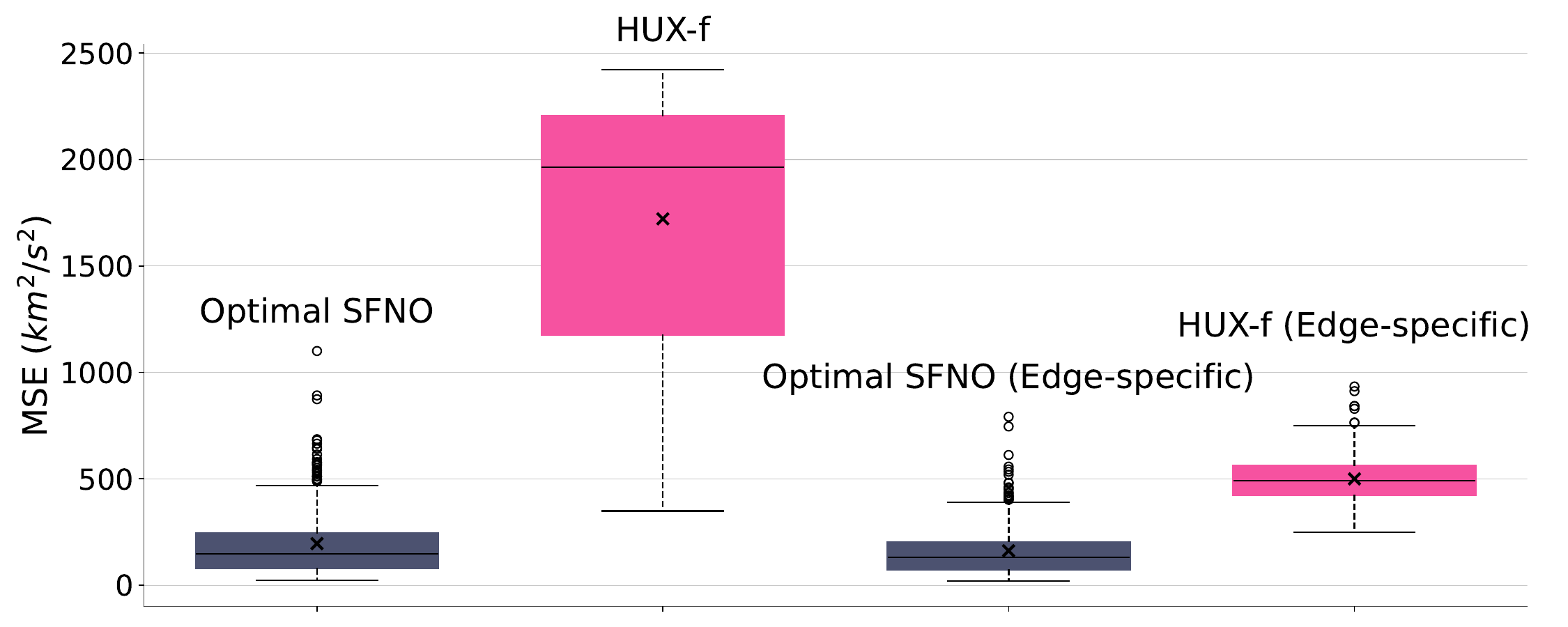}
    \caption{
    Comparison of MSE on the test set between the optimal 8-layer, 256-channel SFNO model and the HUX-f method, evaluated over the full spatial domain and within high-gradient boundary regions.
}
    \label{fig:mse_sfno_hux}
\end{figure}

\begin{figure}[b]
    \centering
    \includegraphics[width=1\linewidth]{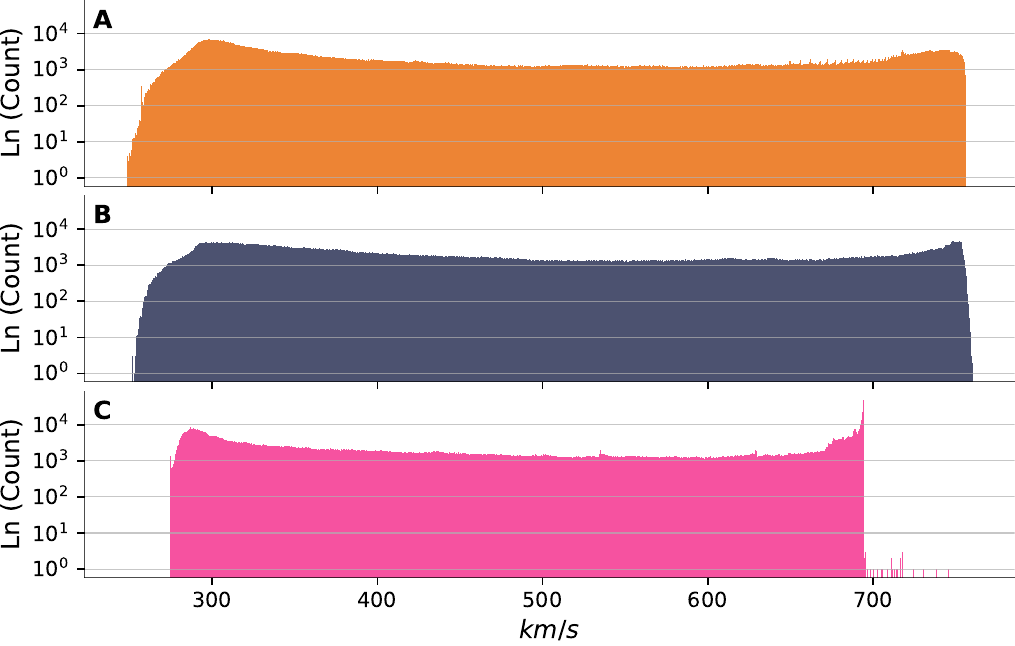}
    \caption{
Distribution of solar wind speed values for Carrington Rotation 2293, shown in log scale. (A) MAS ground truth, (B) optimal 8-layer, 256-channel SFNO model, and (C) HUX-f method. The SFNO distribution closely matches the MAS reference, capturing both the high- and low-speed wind regimes more accurately than HUX-f.
}
    \label{fig:hist}
\end{figure}

The 8-layer, 256-channel SFNO architecture consistently outperforms its counterparts across all cross-validation folds, achieving the lowest mean squared error (MSE) among the configurations evaluated. As shown in fig.~\ref{fig:mse_cv}, this robustness highlights the model’s capacity to capture the underlying spatial-spectral structure of the solar wind radial velocity field more effectively than shallower or narrower SFNO variants (loss curves shown in fig.~\ref{fig:loss}). Based on MSE, the optimal SFNO model outperforms HUX not only in terms of overall error but also within edge-specific regions of the domain, where capturing sharp transitions in solar wind velocity is particularly important. As illustrated in fig.~\ref{fig:mse_sfno_hux}, the SFNO shows consistently lower MSE across both global and high-gradient regions, with tighter interquartile ranges indicating greater reliability.
Fig.~\ref{fig:hist} also compares the distribution of solar wind speeds for Carrington Rotation 2293, the hardest case for SFNO (i.e., maximum MSE), highlighting the SFNO model’s improved alignment with the MAS ground truth over HUX-f.

\begin{figure}[h]
    \centering
    \includegraphics[width=1\linewidth]{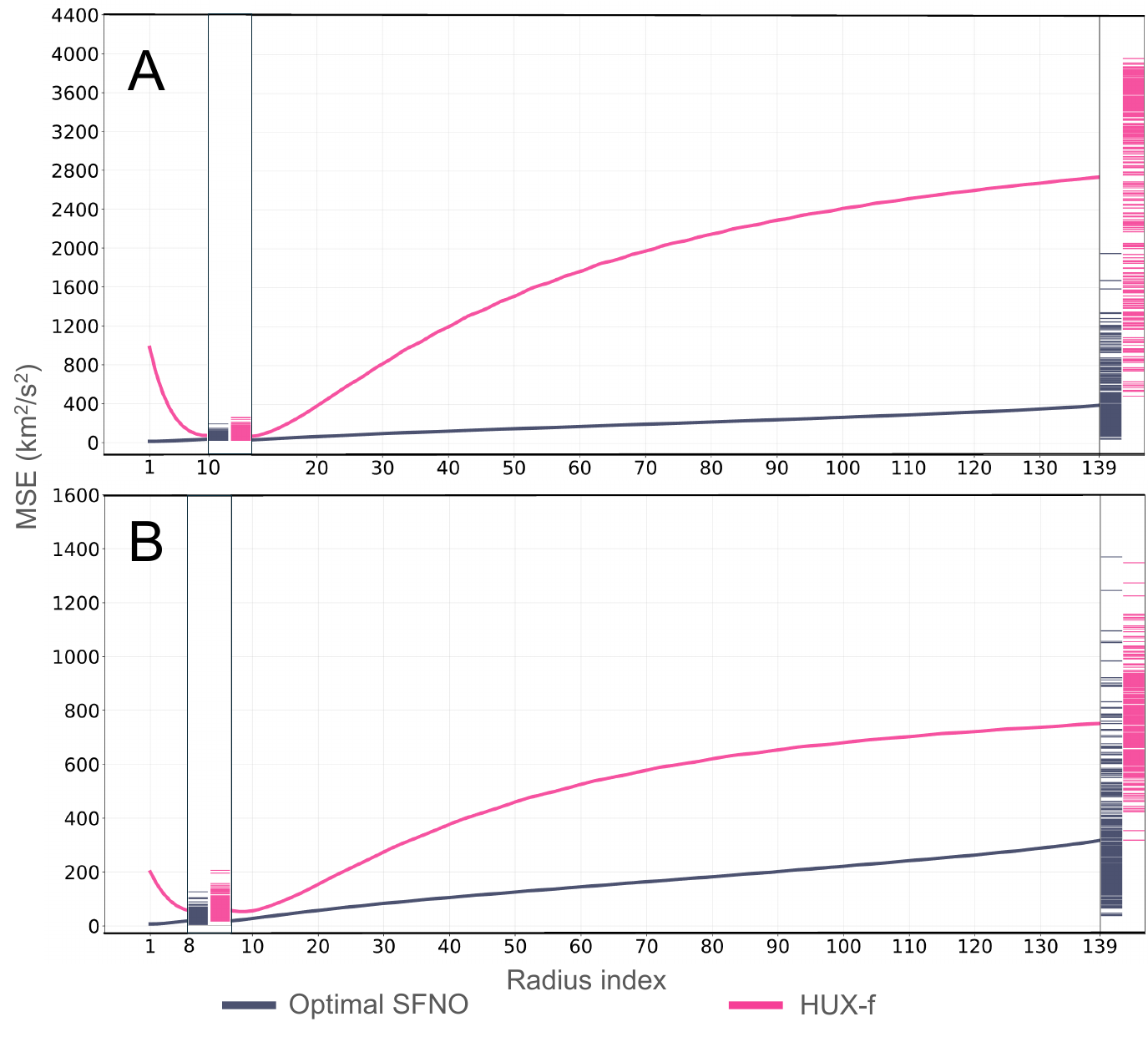}
\caption{
    MSE across radius index (1–139, up to 1~AU) for the SFNO and HUX-f models on the test set. (A) shows MSE over the full spatial domain, while (B) focuses on high-gradient regions. The y-axis indicates the mean MSE per radius, with individual instances for both methods shown at the radius where HUX-f achieves its minimum MSE and at the final radius. In (B), SFNO shows greater spread, aligning more closely with HUX-f in high-gradient areas. The minimum MSE achieved by HUX-f is also indicated for reference.
}
    \label{fig:mse_per_slice}
\end{figure}

Moreover, as shown in fig.~\ref{fig:mse_per_slice}, the MSE distributions for both SFNO and HUX evolve with radial distance, but SFNO consistently shows a lower mean error across all radii. In both the unmasked and masked cases, SFNO’s error distribution is negatively skewed, with most errors clustered at low values. In contrast, HUX’s error distribution is positively skewed, This highlights SFNO’s ability to produce more accurate and stable predictions, particularly in capturing the dominant high-velocity structures in the solar wind.

\begin{figure}
    \centering
    \includegraphics[width=1\linewidth]{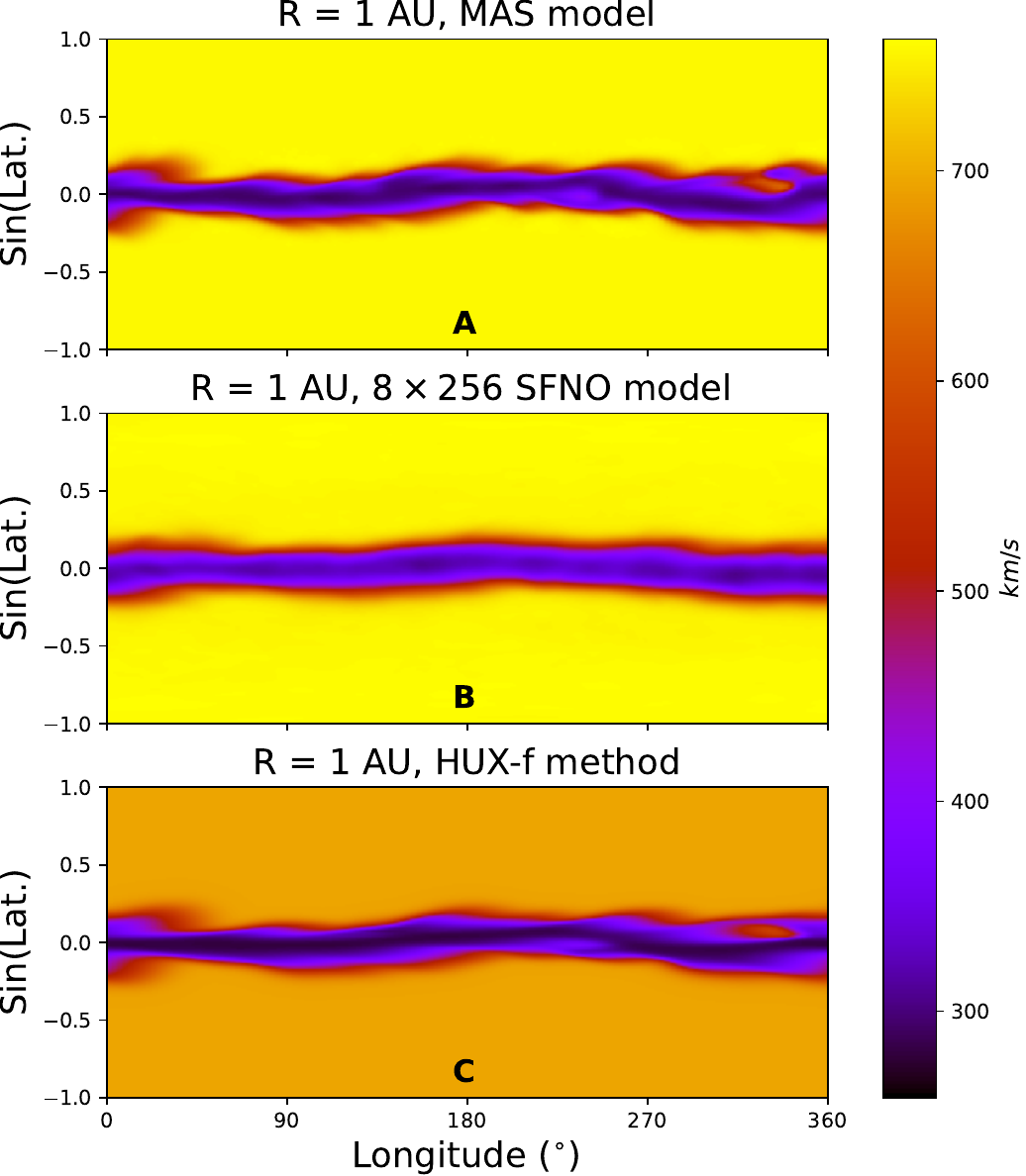}
    \caption{
Solar wind speed at 1~AU for Carrington Rotation 2228. (A) Ground truth from the MHD solution. (B) Estimate from the optimal SFNO model. (C) Estimate from the HUX-f technique. This case represents one of the easiest for SFNO, with low error even in regions of moderate gradients.
}
\label{fig:best}
\end{figure}

\begin{figure}
    \centering
    \includegraphics[width=1\linewidth]{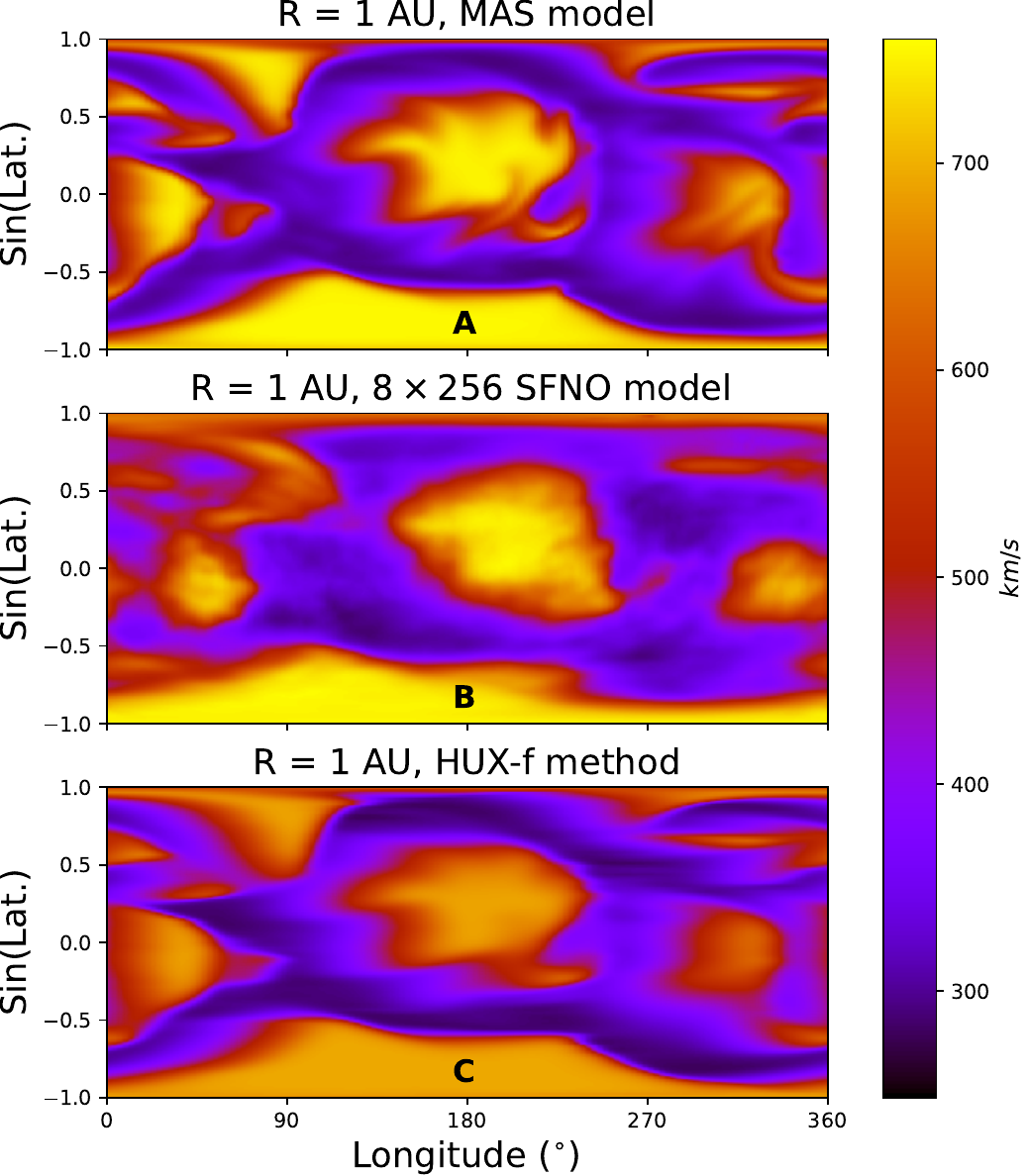}
    \caption{
Solar wind speed at 1~AU for Carrington Rotation 2293. (A) Ground truth from the MHD solution. (B) Estimate from the optimal SFNO model. (C) Estimate from the HUX-f technique. This case is among the most challenging for SFNO in high-gradient regions (MSE).
}
    \label{fig:worst}
\end{figure}

The optimal SFNO shows superior accuracy in both fast and slow wind, with especially strong performance in high-speed polar regions (figs.~\ref{fig:best}, \ref{fig:worst}). Unlike HUX-f, which relies on an empirical acceleration term, SFNO learns directly from high-fidelity data, enabling it to capture polar outflows and the true acceleration profile with greater flexibility and precision. Its spectral representation is particularly well suited for the anisotropic dynamics characteristic of polar regions.

\begin{figure*}[!h]
    \centering
    \includegraphics[width=1\linewidth]{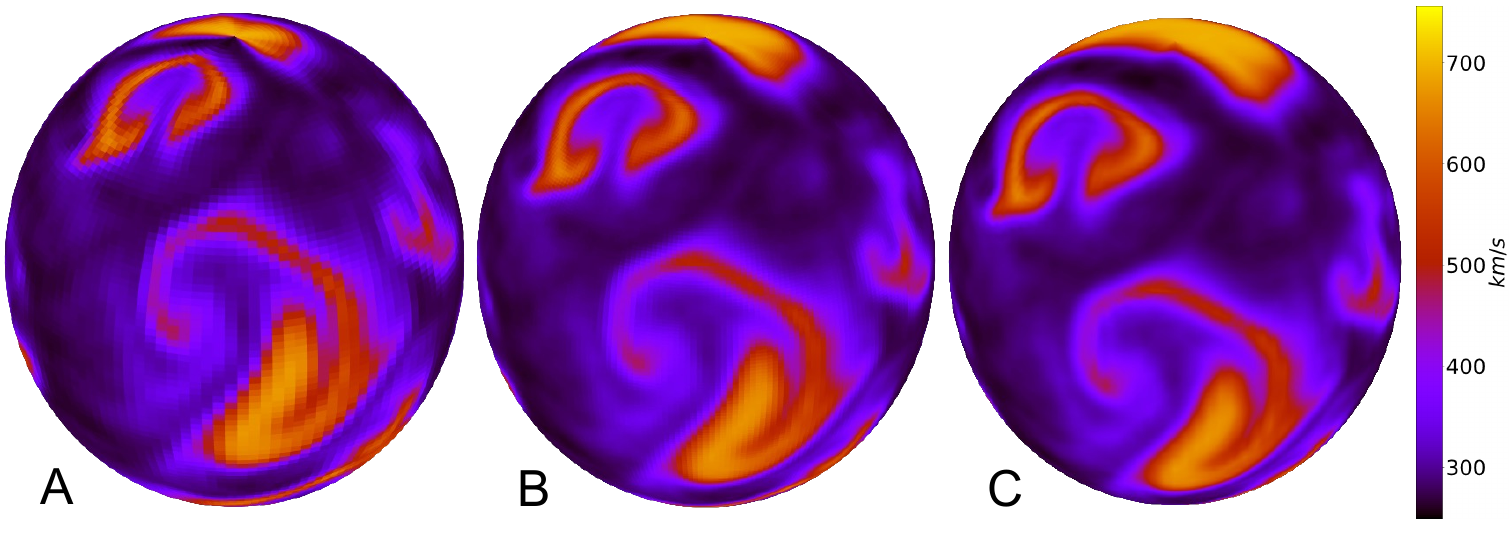}
\caption{
Resolution invariance of the optimal 8-layer, 256-channel SFNO model, which operates in the spectral domain using spherical harmonics. Spherical surfaces of radial velocity predictions at \(\approx45\,R_\odot\) for Carrington Rotation 2293 are shown at three resolutions: (A) the original training resolution \((110, 128)\), (B) doubled resolution \((220, 256)\), and (C) \(4\times\) resolution \((440, 512)\). Higher-resolution outputs are generated using interpolated inner boundary inputs. SFNO maintains consistent spatial structure without re-training.
}
    \label{fig:resolution_invariance}
\end{figure*}

Although SFNO outperforms or matches HUX-f across standard quantitative metrics, as shown in table~\ref{tab:metrics}, visual inspection suggests that HUX-f sometimes produces flow fields that more closely resemble the MAS reference in terms of large-scale structure. This inconsistency highlights a limitation in the current set of metrics, which may fail to capture perceptual or physically meaningful aspects of solar wind morphology. Developing more tailored evaluation criteria that align with expert judgment and domain-specific priorities may be necessary to fully assess model performance.

\begin{table*}[h]
\centering
\caption{
Comparison of SFNO and HUX-f on the test set using various distance and similarity metrics, along with relative change.
}

\resizebox{\textwidth}{!}{
\begin{tabular}{l|l|l|l|l|l|l}
\textbf{Model} & \textbf{MSE $\downarrow$} & \textbf{Edge MSE $\downarrow$} & \textbf{Earth Mover's $\downarrow$} & \textbf{MS-SSIM $\uparrow$} & \textbf{ACC $\uparrow$} & \textbf{PSNR $\uparrow$} \\
\hline
\textbf{SFNO} & 195.591 & 161.993 & 111.0 & 0.9907 & 0.9963 & 35.958 \\
\textbf{HUX-f} & 1,721.4 & 499.534 & 427.976 & 0.9723 & 0.9584 & 25.599 \\
\textbf{Rel. Change (\%)} & 88.63\% & 67.58\% & 74.06\% & 1.89\% & 3.96\% & 40.45\% \\
\end{tabular}
}
\label{tab:metrics}
\end{table*}

Regardless of these differences, it is important to recognize that SFNO is a fully data-driven model, whereas HUX is a physics-based numerical solver. From the perspective of surrogate modeling, SFNO’s ability to match or exceed HUX’s performance without explicitly encoding governing equations demonstrates its value as a fast, flexible alternative. Its efficiency (see Table~\ref{tab:benchmark}) and 
adaptability make it well-suited for downstream tasks such as ensemble forecasting or data assimilation, where rapid evaluations are essential. As shown in Figure~\ref{fig:resolution_invariance}, SFNO also generalizes across spatial resolutions \emph{without re-training}, producing consistent radial velocity structures even when the input boundary is upsampled and evaluated at \(2\times\) and \(4\times\) the original resolution.

\begin{table}[h]
\centering
\caption{
Runtime and memory comparison for a single inference using the optimal SFNO model and HUX-f, run on CPU with identical system settings. SFNO memory includes both model size and inference usage.
}
\begin{tabular}{l|l|l|l}
\textbf{Method} & \textbf{Time (s)} & \textbf{Total Mem. (MB)} & \textbf{Infer. Mem. (MB)} \\
\hline
Optimal SFNO & 1.2982 & 1,233.16 & 258.18 \\
HUX-f        & 6.7577 & 15.39   & 15.39 \\
\end{tabular}
\label{tab:benchmark}
\end{table}

\section*{Conclusion}
Understanding and predicting the solar wind's behavior is essential for safeguarding satellites, communications, and power systems. In this work, we proposed the first data-driven surrogate for solar wind modeling using the Spherical Fourier Neural Operator, showing performance comparable to traditional solvers like HUX. Future directions include developing more customized loss functions, defining skill scores tailored to solar wind dynamics, and incorporating additional variables of the solar wind, such as plasma mass density $\rho$, through physics-informed training to better capture the underlying relationships between solar wind components.

\section*{Acknowledgements}

The data used in this work are publicly available at 
\url{https://predsci.com/data/runs}.
The code implementation can be accessed at \url{https://github.com/rezmansouri/solarwind-sfno-velocity}. Moreover the authors gratefully acknowledge support from NASA’s HSO-Connect program (under grant number 80NSSC20K1285), as well as PSP WISPR contract NNG11EK11I to NRL (under subcontract N00173-19-C-2003 to Predictive Science).


\bibliographystyle{IEEEtran}
\bibliography{refs}

\end{document}